# Predictive Patentomics: Forecasting Innovation Success and Valuation with ChatGPT[1]

**Stephen Yang**

June, 2023

**Abstract**

Analysis of innovation has been fundamentally limited by conventional approaches to broad, structural variables. This paper pushes the boundaries, taking an LLM approach to patent analysis with the groundbreaking ChatGPT technology. OpenAI's state-of-the-art textual embedding accesses complex information about the quality and impact of each invention to power deep learning predictive models. The nuanced embedding drives a 24% incremental improvement in R-squared predicting patent value and clearly isolates the worst and best applications. These models enable a revision of the contemporary Kogan, Papanikolaou, Seru, and Stoffman (2017) valuation of patents by a median deviation of 1.5 times, accounting for potential institutional predictions. Furthermore, the market fails to incorporate timely information about applications; a long-short portfolio based on predicted acceptance rates achieves significant abnormal returns of 3.3% annually. The models provide an opportunity to revolutionize startup and small-firm corporate policy vis-à-vis patenting.

Keywords: AI, ChatGPT, Large Language Model, Machine Learning, Innovation, Patents, Patent Success, Patent Applications, Patent Value, Textual Analysis, Natural Language Processing, FinTech

JEL Classification: G30, O32, O34

---

[1] Contact: Stephen Yang, Pace Academy, Email: stephen.yang25@paceacademy.org

# 1. Introduction

Patents have been the subject of extensive study as the embodiment of innovation and Schumpterian creative destruction. However, existing literature measuring the impact of patents is limited to ex-post analysis. Specifically, a wealth of research examines the value and significance of patents, whether measured through market returns, citations, or other proxies (e.g., Hall, Jaffe, and Trajtenberg, 2005; Kogan, Papanikolaou, Seru, and Stoffman, 2017, henceforth KPSS). Additionally, the technology has not existed to comprehensively analyze the content of patents beyond structural observations such as classifications or firm characteristics. This paper fills the void by creating predictive models for patent value and application acceptance using the revolutionary ChatGPT technology. An ex-ante approach has significant unexplored economic implications for innovating firms and investors.

This paper examines the following research questions. First, is it possible to accurately predict patent value and application acceptance? Second, how can existing models regarding patent value be supplemented and improved using Large Language Model (LLM) technology? That is, can qualitative, non-trivial information regarding the quality and disruptiveness of the innovation be extracted from the application text? Third, can these predictive models help companies and investors make better decisions regarding innovation?

I study a sample of roughly 900,000 applications and 2,000,000 patents, from 2001 to 2020. The deep learning model for patent acceptance starts from the ChatGPT textual embedding, which captures information about the application text into a numerical feature vector. The embedding is supplemented with basic structural variables such as patent classification or firm size and fed into a neural network. LLM embeddings capture many qualitative aspects of textual information previously unavailable to researchers, including but not limited to, sentiment, writing quality, technicality, and so on (Huang, Wang, and Yang, 2022). Given that patent applications are reviewed by human readers, these variables concerning the text should be powerful predictors for the application's ultimate success or failure. The skewed nature of the dataset (as around 70% of applications are eventually granted as patents) makes it difficult to predict acceptance accurately while keeping type 2 error reasonably controlled. A sophisticated deep neural network model is the main predictor and achieves an F1 score of 86%. The performance of this model is significantly superior to a benchmark neural network without the embedding features and an XGBoost approach (a decision-tree-based machine learning model) with the complete set of features. This demonstrates that both the textual embedding and the deep neural network structure are central to the primary model's superior performance. Especially important is that the model performs exceptionally well at predicting the "worst" and "best" applications. The difference in the success rates of the worst and best applications predicted by the primary model and the benchmark model without the textual embedding is economically significant, at around 10%.

Visual examination finds trends of vague and repetitive language in the "worst" predicted applications where structural variables such as firm size and reputation alone would indicate a high chance of success. Additionally, I create a proxy for *application quality* by training separate neural networks only on the ChatGPT embedding, such that the models only have the textual information as inputs. I find that applications filed by larger, older, and more experienced firms all exhibit significantly higher *application quality* and thus experience a substantial advantage in the patenting process. This is consistent with hypotheses that the greater resources and experience available to industry leaders allow them to perfect their applications and apply inside knowledge of USPTO procedures to "game" the process.

I next apply similar approaches to predict the financial value of patents as per KPSS. The prediction uses the same feature variables and neural network design as for acceptance. In this case, the primary model achieves an impressive result of 42% adjusted R-squared for the full model. This result is a substantial improvement of 24% over a benchmark model trained without the embedding. Additionally, an XGBoost model trained for comparison on the complete feature set still outperforms the benchmark neural network by 10%. Together, these results prove that the ChatGPT textual embedding provides economically significant, previously inaccessible information. These results are achieved partially through innovative approaches to network design and dependent transformation, which are discussed further in the body of the paper.

These new models have significant potential economic implications. First, I use the predictive model for application acceptance to revise the KPSS estimation of patent value. KPSS introduces a scaling factor $1/(1-p)$ that accounts for the fact that investors expect a patent to be accepted with a probability $p$. They use a constant $p$ for all patents, based on the argument that it is challenging to predict patent acceptance. I propose replacing the blanket constant scaling factor with a factor calculated from the $p$ given by the machine learning model in this paper. Since the prediction from the embedding can simulate the analysis of human experts or even natural language processing models employed by investors, the revised factor enables a potentially more realistic valuation of patents. The median proportional deviation of the new measure from KPSS is 1.46 times, and the mean deviation is 2.65 times. The most substantial contribution of this estimator is in the outlier cases, i.e., very strong or weak applications for which the market reaction will be otherwise proportionally under or overestimated. Moreover, if the market does accurately judge the application's chances of success and acts efficiently, the potential undervaluation of other methods is nearly unbounded: incredibly disruptive and impactful inventions may be evaluated by the market as being near certain grants, and thus the reaction following acceptance will be extremely undersized. For the patents deemed "best" by the predictive model, the assumption of a constant acceptance rate may undervalue as much as tenfold. However, it is worth mentioning that the average acceptance rate in my sample is 72.4%, higher than the 55% in KPSS's larger sample. Adjusting the KPSS scaling to this acceptance rate, the median deviation in the valuation is still a sizable 51%, and the mean is 1.24 times.

Second, the models have the potential to enhance the patenting policies of companies substantially. When companies' applications have a low predicted probability of acceptance based on the predictive model and are later accepted by the USPTO, there tend to be meaningful changes between the grant and application texts that significantly improve the predicted chance of success. However, while our findings are significant at the 1% level, the subsample size is small, indicating that this strategy is severely underutilized. This trend suggests that companies, especially startups and small firms that lack the expertise and resources to effectively "game" the system, can improve their innovation process by using machine learning models to screen and fine-tune applications before sending them to the USPTO. The resulting outcome will be higher acceptance rates and more impactful innovation by these firms. This process could encourage innovation and R&D investment, as the otherwise nebulous risk of failed applications threatens the loss of capital and revealing the new technology to competitors.

Third, the models provide access to new information with regards to patent applications, previously not leveraged by the market. I construct a firm-level measure of *application strength* for firms, based on the number of applications and average predicted chance of success published in a certain month. This provides a good proxy for both the innovative strength of a firm and the latent patent value. The second of these may not have been efficiently incorporated by investors, as most news coverage focuses on patent grants rather than applications. Indeed, a long-short portfolio constructed based on *application strength* achieves statistically significant yearly abnormal returns of 3.3%. This identification is further supported by the evidence that portfolios constructed based on measures from granted patents fail to achieve abnormal returns.

Our study contributes to several strands of literature. First are the studies of the valuation of innovation. Determining the value of innovation is vital to both the market and individual firms, as innovation has an outsized impact both socially and for growth (Drucker, 2014; Bloom, Schankerman, and Van Reenen, 2013; KPSS, 2017). There is a long history of literature studying patent valuation. Some of the earliest works directly examined the relationship between patent release and stock returns, finding evidence that patent grants are highly correlated with market returns (Pakes, 1985; Austin, 1993). Hall, Jaffe, and Trajtenberg (2005) use raw patent citation count instead as a proxy for innovative success and find a strong correlation with the firm's market valuation. KPSS develop a new measure based on market reaction to the announcement of patent success. This paper builds on the KPSS value, scaling based on the prediction of application success, which better proxies the actual investor evaluation of the innovation. This method has the potential to help companies and investors to recognize better and distinguish the most impactful innovations, a known problem (Cohen, Diether, and Malloy, 2013). Kelly et al. (2021) also take a machine learning approach to patent evaluation but are instead measuring technological disruptiveness. This is correlated with, but not a direct proxy for, the economic value of a patent (as other factors such as creative destruction and commercialization often take a more direct role than genuine scientific inventiveness) that this paper focuses on.

The second strand of research studies is textual machine learning applications in finance. For the last decade, researchers have used methods from Word2Vec up to BERT in many situations (Hanley and Hoberg, 2019; Li, Mai, Shen, and Yan, 2021; Acikalin, Caskurlu, Hoberg, and Phillips, 2022; Cao, Jiang, Yang, and Zhang, 2023). With the introduction of ChatGPT and related large language models, a few studies have begun implementing this new wave of LLM technology (Lopez-Lira and Tang, 2023; Hansen and Zazinnik, 2023). This paper applies the state-of-the-art embedding technology by OpenAI's ChatGPT, which encodes richer information and higher-level concepts than previous models. The superior predictions achieved by the models indicate that the development of sophisticated LLMs has far-ranging implications for finance more broadly.

## 2. Data, Sample, and Variables

### *2.1 Data Sources and Sample*

The primary data source is the USPTO's PatentsView service, which provides complete documentation for all patents from 1976 and applications from 2001 in the United States. I also use the USPTO Bulk Data Storage Service (BDSS) in order to obtain the full text of some patents and applications. This is supplemented by market information from CRSP/Compustat provided by the WRDS service. I use the entire sample of patents and applications, contingent on the assignee having a valid "permno" to link to market data. Observations with multiple assignees are dropped to avoid duplication. The final sample contains 855,891 applications and 3,177,942 granted patents. The disparity between the two is partly a result of the USPTO not publishing application texts prior to 2001. I thus truncate the patent dataset, in most tests, at 2001, reducing the sample size to 2,239,148. The remaining difference is primarily a result of applications filed before 2001 as well as missing assignee data in the application files, i.e., the USPTO data lacks valid company assignees for a large portion of applications. Additionally, I use a rolling window training sample; all models are trained on data from the three preceding years, i.e., the predictive model for application success in 2004 is trained on all applications from 2001 to 2003. This ensures that all trends captured by the model will still be contextually relevant and avoids the problem of using unavailable information for prediction, all while maintaining a sufficiently large training sample for deep learning techniques. Finally, the application data is truncated by nature, as evaluation can often take several years. Thus both 2021 and 2022 are excluded from the application sample. In 2020, the truncation effect is less prominent, but the pandemic also introduced new trends not present in training data, so it is also excluded. As a result, the effective test sample for applications is from 2004 to 2019.

*2.2 Variables*

Our primary independent variable is the embedding vector generated by the GPT model, obtained from OpenAI's publicly available API. The title and abstract of all applications and patents are combined and fed through the model to obtain the embeddings. Each embedding vector is a list of 1,536 numbers, which contains many dimensions of information about the text (e.g., sentiment, topic, et cetera). Fundamental structural variables supplement this, which is critical to provide context, such as firm size and industry, to any given application or patent. Specifically, I use the CPC classification system (i.e., Class A, B, C, and so on), a USPTO-generated measure of "AI" patents, separate classes of ICT, biotechnology, and high tech, which I generate based on the classification by the European Union (2006), number of CPC classes, the natural logarithm of the number of claims, whether an assignee is a research institution, and the Fama-French 12 industry classification, obtained from Kenneth French's website. For application prediction, the natural logarithm of market capitalization is also included. The market capitalization is calculated in the nearest quarter prior to publication and adjusted for inflation.

The dependent for success prediction is a simple binary dummy, equal to one if the patent is granted and zero if it is rejected. I use the KPSS measure for patent value, calculated based on the market reaction to the announcement of patent acceptance in a three-day window per the procedure in the 2017 paper.

## 3. Model

### *3.1 Neural Network*

The availability of OpenAI's ChatGPT API to the general public, specifically the release of the Ada-002 LLM embedding model on December 15, 2022, gives researchers unprecedented access to deep-learning capabilities. Ada-002 is OpenAI's top-of-the-line embedding model, capturing as much as four times the context of previous models, and provides an opportunity to utilize nuances of textual information previously inaccessible or impractical due to labor constraints. This technology is a significant improvement, as previous methods such as Word2Vec fail to account for context, thus losing the meaning from phrases or conditional modifiers as well as only being useful for basic tasks such as sentiment analysis which do not require recognition of higher-level ideas. Ada-002 is a transformer model, which means it uses "attention" vectors to ultimately output an embedding that transforms an entire body of text into a vector of 1536 real numbers. These methods capture the meaning and nature of the text in its entirety rather than in disjoint parts.

This embedding vector then has many applications, from being used in a generative model such as ChatGPT relies on to classification (such as review sentiment analysis, a popular benchmark in the neural network community) and even unsupervised clustering problems. This paper uses the vector as a feature (with 1536 components) in a secondary machine learning approach to predicting both the value of patents and the success of applications, with powerful results. Given that previous literature regarding patents utilizes GLoVE or Word2Vec, and in finance generally, only BERT or ELMO, this is a substantial step forward and provides potential for application to other fields. It is likely possible to achieve even better results, particularly for value, by including more variables from the corporate side, as the models in this paper only utilize fundamental variables such as firm size and patent class in order to demonstrate better the significance and predictive power of the embedding features.

For both predictive models, I feed the embedding vector with the structural variables as the feature vector into a three-layer feed-forward neural network, or multi-layer perceptron (MLP).[2] The choice of activation function is often overlooked in economic applications of neural networks. However, the choice of function is in fact perhaps the most crucial hyperparameter (besides layer design). Simple *ReLU* is the most used activation function, which equals $0$ if $x$ is negative, and $x$ otherwise. Some other traditional functions are *tanh*, *softplus* ($\ln(1 + e^x)$), and *ELU* ($e^x - 1$ for negative $x$, and $x$ for positive $x$). I utilize a custom activation function of "*Mish*," defined as such:

$$Mish(x) = x \cdot \tanh(\ln(1 + e^x)) \qquad (1)$$

*Mish* is a revolutionary addition to the traditional family of activation functions. I find *Mish* to perform significantly better in cases with large sample sizes and feature count, likely in order to capture highly non-linear and irregular relations between the embedding features and because of its ability to handle negative inputs, while *ELU* or even *ReLU* can perform similarly in cases of low dimensionality, such as when excluding the embedding vector. The invention of the *Mish* formula is a major innovation from 2020 and provides a significant improvement in performance over traditional *ReLU* activation (Misra 2020). Several of these activation functions are graphed in Figure 1. *Swish* is another contemporary activation function, defined as $x \cdot sigmoid(\beta x)$, where $\beta$ is learned (Ramachandran, Zoph, and Le, 2017). Although *Mish* and *Swish* appear similar, there are differences in their respective first and second derivatives which change how the models evolve (in my testing, *Swish* fails to match even *ReLU*). More implementation details are provided in Appendix A.

---

[2] All programming is done in Python. The neural networks are built and trained using TensorFlow's Keras package.

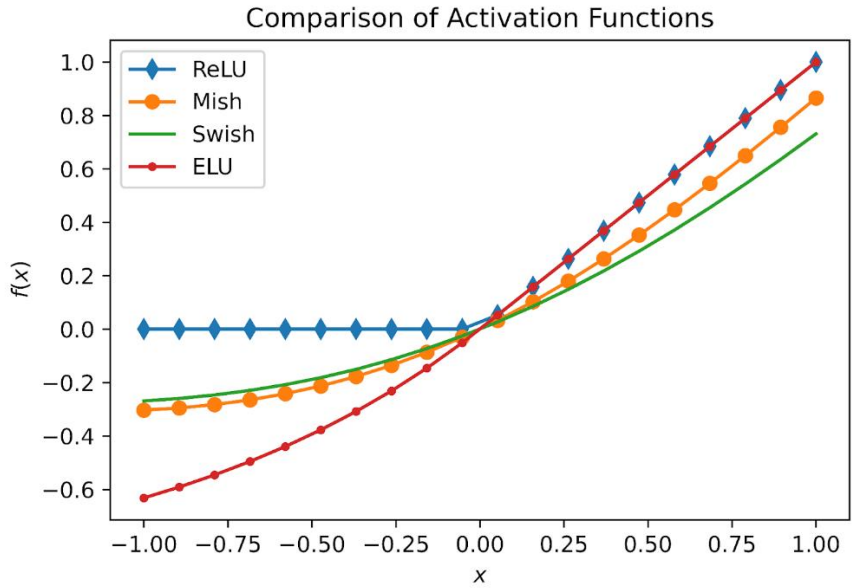

Figure 1: Comparison of Activation Functions

*3.2 Predicting Acceptance*

These methods are first applied to predict the likelihood of application acceptance. I treat this as a binary classification problem, with the dependent variable *Accepted* equal to 1 if the application becomes a granted patent and 0 otherwise. Three cases are tested. First, the primary deep learning model is given the complete list of features, excluding a few variables that the USPTO does not provide in processed application data on PatentView (number of claims, AI classification). The dependent *Accepted* is heavily skewed by nature, as the USPTO accepts roughly 70% of patent applications. Thus, the model is optimized for binary cross-entropy (log-loss) to best adapt to the skewness.[3] Second, I benchmark by training a similar deep learning model without the embedding features to determine the incremental improvement generated from the textual information. Third, I test XGBoost (a commonly used machine learning method relying on decision trees) on the comprehensive set of variables for comparison. In untabulated results, a linear SVC classifier is trained, which fails to beat the null predictor (predicting success for every application), potentially due to the high dimensionality of the features.

---

[3] Optimizing for Accuracy or F1 Score biases the model towards recreating the null predictor.

Five statistics, commonly employed for the evaluation of binary classification models, are presented in Table 1, illustrating the year-by-year performance of the primary deep learning model. AUC is the area under the Receiver Operating Characteristic (ROC) curve and measures the probability of ranking a positive instance higher than a negative instance. AUC is the best criterion for evaluation due to the aforementioned skewness of the sample. Accuracy is the overall percentage of correct predictions. Precision is the percentage of true positives in all labeled positives, and Recall is the percentage of true positives in all real positives. F1 Score is the harmonic mean of Precision and Recall. The model performs well in all measures, with an average Accuracy of 77%, and an average F1 Score of 86%. Importantly, it achieves an AUC score of 57%, a Precision of 79%, and a Recall of 95%, averaged across all years.

[Insert Table 1]

Second, Table 2 shows the comparison of the main model with benchmarks. While the benchmarks achieve deceptively similar results in Accuracy and F1 score, both fall short in AUC, demonstrating that the conjunction of the ChatGPT embedding and advanced deep learning techniques is the key to distinguishing the bad applications from the mediocre or good. This is again demonstrated by the excessively high Recall (true positives correctly classified) of both benchmarks. Nearing 100% Recall suggests that both models approach the null predictor, as it is obviously otherwise impossible to avoid false negatives completely.

[Insert Table 2]

I qualitatively demonstrate this difference in AUC between the principal model and the benchmark without the embedding features by evaluating the predicted "worst" and "best" applications from each year in sample for both models. The results are presented in Table 3. The fully trained model performs exceptionally well at both the highest and lowest end of prediction. For example, the yearly top 100 predicted applications have a 96% acceptance rate, and the bottom 100 only 26%. Even enlarging the yearly cutoff to 1000, the model still maintains a success rate of 94% for top applications and 34% for the bottom. The ability to accurately predict the worst applications is surprising because the data is skewed so heavily toward acceptance. In comparison, the benchmark reaches a comparatively weak 87% rate for the top 100 yearly predictions and 37% for the bottom 100. Clearly, the primary model is significantly better at picking out applications likely to succeed and those likely to be denied. Thus, the benchmark is significantly less valuable economically, as it neither provides the ability to accurately distinguish the best applications to invest in, nor the worst to screen out; these are only possible with the additional textual information provided by the embedding.

[Insert Table 3]

Although there are no consistent, generally agreed-upon methods for interpreting neural network predictions en masse, visual investigation finds consistent trends in some of the worst predictions in sample, which the ChatGPT textual embedding potentially isolates for the predictive model to use. Appendix B shows the title and abstract of a Class A (human necessities) patent application filed by a leader in the pharmaceutical industry. Given these structural variables, human analysts would likely predict a high chance of success. It is from a market leader in the drug industry applying for a patent for a newly developed drug. However, the model is highly critical and correctly predicts that it will not be accepted, assigning a meager chance of 5.9%. This indicates that it must have extracted additional information from the text of the abstract and title. Reading the application without a Ph.D. in organic chemistry, it is unclear what the purpose or innovation of the invention is (a new synthesis, delivery method, et cetera). Notably, the abstract repeats the same sentence twice ("the present invention relates to pharmaceutical compositions…") with a change in a chemical name, a perplexing choice that makes the application nearly unreadable. The predictive model likely accounts for the subpar, unclear writing in its assessment of the application.

Additionally, I investigate whether economic factors drive these trends of worse application writing. For example, do older and more experienced firms submit more polished applications, given their detailed knowledge of USPTO processes? In order to answer this question, I first isolate the effect of the embedding on firm success by training a distinct neural network with only the embedding as input. The resulting prediction of application success proxies directly for *application quality*, as the model only factors in the text itself. I then conduct the following regression, for firm-application-time $(i, j, t)$:

$$Application\ Quality_{i,j,t} = \beta_1 Size_{i,t} + \beta_2 Age_{i,t} + \beta_3 Application\ Stock_{i,t} \\ + (\gamma \cdot Controls_{j,t}) + \delta_i + \epsilon_{i,j,t} \tag{2}$$

Firm size is the natural log of shares outstanding times price, adjusted for inflation, and age is the number of years since the firm first appeared in CRSP. Application stock is defined as the cumulative number of patent applications (regardless of success) filed by a firm before the month of the current application. The results of the regression are shown in Table 4. Additional specifications are also tested with each independent alone with firm-fixed effects. Note that year-fixed effects are not suitable for this situation, as combined with firm-fixed effects, the controls would completely absorb firm age. Using patent-level controls, I find strong positive correlation to *application quality* from all three independents, firm size, age, and application stock, all significant at the 1% level. The results are robust when removing patent controls and testing each independent separately. Overall, this demonstrates that a firm's experience and resources significantly affect the quality of its applications.

[Insert Table 4]

The prediction of application success has multiple potential applications. First, firms with access to LLM technology can "screen" patent applications before submission and revise them to obtain the highest chance of success. This process could encourage greater investment in R&D, as risk-averse managers can be more confident. Second, I propose the use of a predictive model of acceptance to develop a better measure of genuine patent value. More details are found in Section 4.

*3.3 Predicting Value*

Next, I predict the KPSS measure of economic value, determined via market response to the publication.[4] The same procedure as acceptance is followed to create the sample, training with a rolling window, and the dependent is first calculated via the KPSS procedure and then scaled by market capitalization (again, calculated in the quarter prior to publication and adjusted for inflation). However, direct training fails to produce good results, as value is highly skewed by nature even after the adjustment. As machine learning techniques work best on demeaned dependents with lower variance and skewness, I transform the KPSS value before testing the models. I apply several normalization methods in pre-processing, including BoxCox, quantile normal, $ln(1 + y)$, and simple z-score normalization (standardization). Of these, BoxCox, a method introduced in Box and Cox (1964), performs the best. This method is commonly used in the financial literature to perform "BoxCox Regression," in which OLS is applied to a dependent which has been put through the BoxCox transformation (e.g., Bhagat and Frost, 1986). In cases of highly skewed positively valued dependent variables, applying BoxCox significantly improves the performance of OLS. However, machine learning papers in finance and economics generally only apply $ln(1 + y)$, standardization, or no transformation to the dependent when training a neural network, while social sciences more broadly accept BoxCox as the historically proven best practice for resolving normality and heteroscedasticity concerns (Osborne 2019). The application of BoxCox in the models is thus an innovation as it significantly improves predictive power in terms of R-squared while still maintaining applicability as relative ranks are unchanged. The transformation is calculated as follows:

$$y_i^{(\lambda)} = \begin{cases} \frac{y_i^\lambda - 1}{\lambda}, & \text{if } \lambda \neq 0, \\ \ln(y_i), & \text{if } \lambda = 0. \end{cases} \qquad (3)$$

---

[4] The textual embedding is, by nature, directly correlated with the revised valuation and thus would be an overly powerful predictor. Any tests performed on the revised valuation would not provide a good metric for objective performance or significance of the embedding.

Note that the use of BoxCox calculates $\lambda$ in sample (traditionally, to maximize log-likelihood), which could invalidate statistical results. I thus calculate $\lambda$ only on the training set and apply the same transformation to the test sample, which has no discernible effect compared to applying the BoxCox transformation to the entire sample. The observed stability of the transformation is expected given the large sample size, such that the data likely approaches the actual distribution of KPSS value. With the transformed KPSS value as the dependent, I again benchmark the model without embedding, then test XGBoost for comparison. OLS and similar linear methods are unsuitable for this situation, because of the high $p$ count and complex interactions within the embedding.

The results of the principal model prediction are reported in Table 5. It achieves a strong adjusted R-squared score of 42%, with a trend of increasing performance with time from 37% in 2004 to 48% in 2020. The improvement is potentially a result of better market incorporation of patent analysis in recent years, fueled by institutional investors' use of machine learning techniques. If true, these trends would mean that the market better accounts for the genuine value of each individual patent.

[Insert Table 5]

Second, the primary model is compared with both benchmarks. Table 6 presents the results of the comparison. First, comparing the full and no embedding models, there is a significant 25% improvement in adjusted R-squared when adding the embedding to the feature matrix, even using the best machine learning methods for both tests. The considerable increase proves that the ChatGPT-based textual embedding vector is a powerful and non-trivial predictor for value even when paired with conventional structural variables. Second, the XGBoost model is unable to match the neural network in performance, reaching only 27% adjusted R-squared with the full list of features. This is 16% weaker than the full model, and the disparity demonstrates the superiority of the advanced machine learning techniques over the XGBoost model. However, even XGBoost still outperforms the advanced neural network trained without the embedding by 10%, further indicating the predictive power and importance of the textual embedding. Thus, the conjunction of both is most important, although the embedding roughly seems more powerful.

[Insert Table 6]

**4. Economic Applications**

*4.1 Screening*

The Supreme Court wrote in 1892 that "the specification and claims of a patent… constitute one of the most difficult legal instruments to draw with accuracy" (Topliff v. Topliff, 1892). As a result, the proposed application "screening" process is potentially of great economic importance. Given a good interpreter of applications and an effective predictive model, innovating firms should be able to maximize their chances of patent success by running prospective applications through the model before submission. This augmentation is of outsized importance to exceptionally small firms and start-ups, for whom the company's survival may very well hinge on the acceptance of a single key patent. Without the resources to hire patent lawyers that major players like Apple employ on a daily basis, these companies often produce weaker application texts and lack inside knowledge of how to "game" USPTO procedures. Specifically, Section 3.2 reports our findings that, controlling for fixed effects, smaller and less experienced (proxied through application stock and firm age) firms tend to produce significantly worse application texts. A good predictive model could thus enable these firms to increase their expected profits significantly while reducing risk for almost no additional cost. In fact, it is documented that patent success plays a prominent role in VC decision making, making the process even more important for start-ups in a highly competitive environment (Häussler, Harhoff, and Müller, 2012). However, the utility of this application is not limited solely to start-ups and firms with limited capital. Larger firms often have patents slip through the metaphorical cracks, likely due to the copious number of inventors working simultaneously. For example, some of the "worst" application abstracts in the sample come from innovation heavyweights such as Pfizer and Ford.

    If this method of screening is feasible, there should be examples in the sample where originally lackluster applications are revised and ultimately accepted with significantly improved patent texts. Thus, I run a test on all application data to search for initially "bad" applications that were turned into "good" patents and accepted. The 500 worst applications from each year in terms of predicted success are gathered and filtered based on ultimate success as patents. The model performs exceptionally well at identifying the "worst" texts and only 2,700 of these roughly 10,000 applications were accepted. Of these, I screen for changes in the abstracts. Rather than crude

approaches such as checking for literal equality or the number of characters changed, I instead again utilize the ChatGPT textual embeddings, this time to measure change. A major change in the abstract is defined as having a cosine distance of at least 0.05 between the embeddings of the application and corresponding patent abstract. This provides a sophisticated measure of similarity between texts, capturing changes in meaning, writing style, sentiment, and more. With the added condition of non-trivial changes in the abstract, there is a group of 42 applications throughout the entire sample. I then re-run the prediction model on the patent grant embedding.

Table 7 reports both predictions of success rate and the "improvement" from application to patent. I find a mean increase of 10.3% and a median increase of 4.1% in the predicted chance of acceptance, with several applications improving by as much as 50% or more. Both the mean and median results are statistically significant at the 1% level. Even lowering the minimum cosine distance to .02, the documented increase in "quality" is still significant at 1%. Also of note is that the average prediction for the application subsample is only 17.4%, significantly lower than the average of 19.8% in the entire group of 2,700, indicating that the worst patents are changed the most. The results shows that revising application texts for a better chance of success is a viable strategy but has likely been underutilized in the past, given the small subsample.

[Insert Table 7]

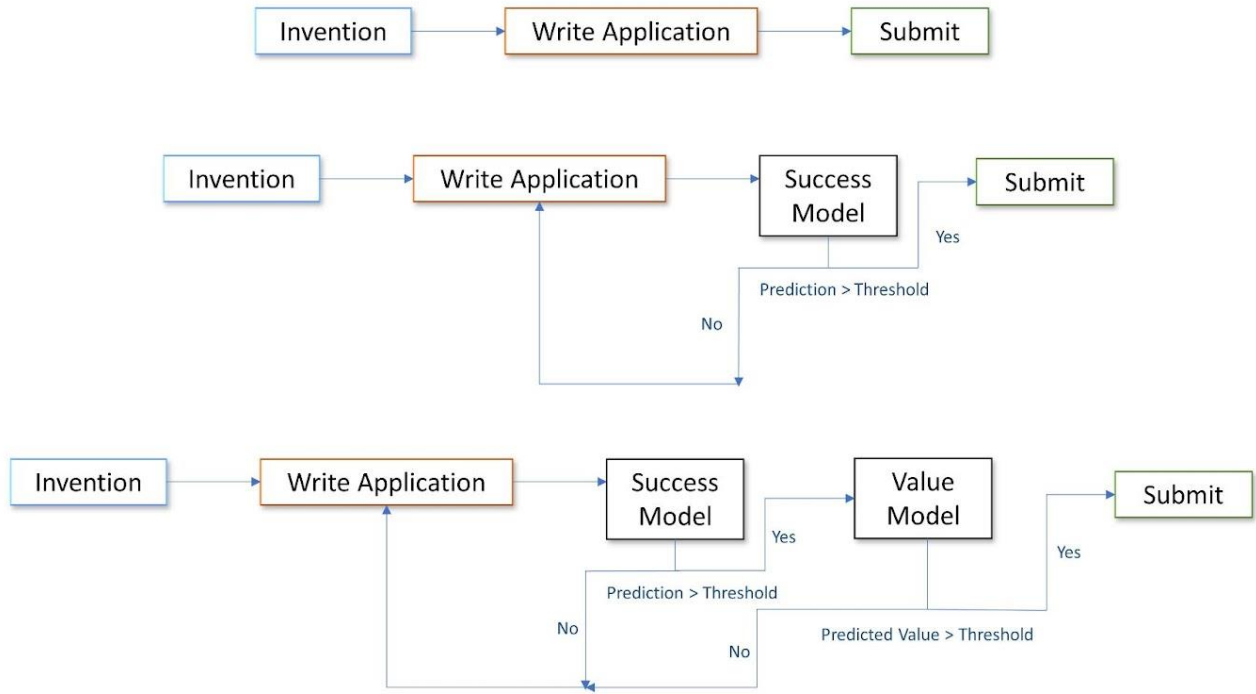

Figure 2: Potential Process of Application Screening

Figure 2 visualizes the decision-making process of firms applying screening. Currently, firms use the straightforward first method, where inventors write an application and directly submit it to the USPTO after finalizing an invention. I theorize an intermediate step in which the firm can repeatedly revise the application text without submitting it to the USPTO using the model until a satisfactory threshold in acceptance chance is reached. Additionally, firms can add a potential second "layer" of screening using the value prediction model. Necessary because the acceptance prediction is not correlated with the value prediction (the USPTO does not accept applications based on estimated economic significance, but rather for originality and non-obviousness), the application of both models could allow firms to maximize both chance of application success and the positive market reaction. I do not find the existence of such trends in the data, likely because existing patent lawyers and experts focus on application acceptance when revising texts. With the introduction of LLM technology to the market, it is increasingly important for firms to consider the reaction of institutional investors to textual information, as has been documented in the cases of corporate disclosure and the news (Cao, Jiang, Yang, and Zhang, 2020; Huang, Tan, and Wermers, 2020). This "feedback effect" should play an outsized role as patent texts are available for public access in bulk data, meaning that firms must monitor the impression they make on the models employed by investors. If nothing else, the use of machine learning models can allow firms to pre-emptively adjust for market reactions.

*4.2 AI Adjusted Patent Value*

Additionally, I propose an adjusted measure of the economic value of patents using the predictive model of acceptance. KPSS scales market reaction (minus variance) by a constant patent acceptance rate, as the returns measured in a three-day window around acceptance are a reaction only to the roughly 45% chance of denial and not the entire patent. This approach does not account for variation in acceptance rates across significant structural variables such as year, firm size, and industry, as it assumes that the market is unable to predict acceptance. However, institutional investors have long had the capability to read patent applications through brute force application of manpower or even just to use structural variables to predict acceptance. As a result, the market likely already at least partially reflect investors' beliefs about the probability of acceptance, especially because large investors have begun implementing machine learning technology in recent years. Thus, future value estimations ought to consider the predictions of the market. For example, if models used by large investors calculate a 95% chance of acceptance for a highly impactful patent, the multiplier is twenty (the value is scaled by $1/(1-.95)$). In this example, the previous model assigns the blanket multiplier of $1/(1-.55)$ and thus substantially underestimates the patent's true value by failing to scale sufficiently.

I test the deviation of the alternative measure from KPSS and report the results in Table 8, compared both to the KPSS assumption of a 55% acceptance rate and an adjusted measure using 72.4% acceptance, which is the acceptance rate in this paper's final, smaller sample. The adjusted measure accounts for some of the disparity in valuation but not all. My scaling factor has a mean proportional difference of 3.65 times and median of 2.46 times from the original KPSS scaling factor; the differences are 2.24 times and 1.51 times respectively from the adjusted factor. There is an outsized difference in valuation because of the aforementioned undervaluation by the constant factor in the cases of the "best" patents. As our measure is prone to outliers, we winsorize the scaling factor and the ultimate valuation at the 1% level. On average, our patent valuation is 22 million dollars larger than the original KPSS value and 17 million larger than the adjusted value. The difference is substantial; for reference, the average KPSS valuation is only 9 million dollars, and 14 million when adjusted.

[Insert Table 8]

### *4.3 Portfolio*

The market potentially underutilizes information about patent applications relative to patent grants. Anecdotally, almost all mainstream news reports regarding innovation focus on fully granted inventions rather than pending applications. Detailed information about applications is only available from USPTO Bulk Data, which many investors may not mine for data. Denied applications can theoretically be of extreme negative value to firms, as they lead to failure in commercialization and, more significantly, allow competitors to "steal" the idea. Pre-emptive knowledge of successful innovation through the "best" applications is also highly valuable.

I thus test a trading strategy based on the predictive models. First, the firm-level measure is defined as the mean predicted chance of acceptance for all patent applications published by a firm in a given month, multiplied by the square root of the number of applications in that month. The measure is referred to henceforth as *application strength*. Each month, the firms are sorted into two groups, above and below the median, based on *application strength*. A traditional long-short portfolio is then constructed for the above and below median groups, holding for a one-month horizon. The abnormal returns are calculated based on the Fama-French three, four, and five-factor models for robustness (Fama and French, 1993, 2015; Carhart, 1997).

Table 9 reports the alphas of the portfolio. The abnormal returns of the long and short portfolios are independently significant at the 10% level. More importantly, the difference in Fama-French four-factor adjusted returns, or overall performance of the long-short portfolio, is 3.3% annually. This result is statistically significant at the 5% level. The returns adjusted by the three and five-factor models are also significant and similar in magnitude. Overall, the evidence suggests that the market fails to sufficiently react to patent applications which provide important information about the innovative strength and latent patent value of firms. In untabulated results, portfolios constructed based on similar measures calculated using granted patents are unable to achieve abnormal returns, further implying that the market does not monitor applications in particular as closely as patents.

[Insert Table 9]

## 5. Concluding Remarks

This paper adopts a new approach to patent analysis with the groundbreaking ChatGPT technology, moving beyond conventional structural variables. OpenAI's state-of-the-art textual embedding allows deep learning models to interpret previously inaccessible information about the impact of each distinct invention. I show a 24% incremental improvement in R-squared predicting patent value when adding the embedding vector. The full model for predicting acceptance is able to clearly isolate the worst and best applications with a roughly 10% improvement in both over the benchmark, which is critical in economic contexts. I propose a revision of the widely accepted Kogan, Papanikolaou, Seru, and Stoffman (2017) valuation of patents. My measure has a mean deviation of 22 million dollars, or 2.5 times from the KPSS measure and accounts for institutional analysis of patents by providing an alternative to the hypothesis of uniform acceptance rate. Additionally, the models provide an opportunity to firms, especially startups or small firms, to enhance their patenting and innovation processes. Finally, a long-short portfolio constructed based on the strength of applications filed by firms achieves significant annual returns of 3.3%, indicating that the information captured by my models has, as yet, not been efficiently incorporated into the market.

This paper also leaves room in the future for further research alongside the development of machine learning technology. First, LLMs currently need to be better equipped for even basic symbolic processing (as can be seen when asking ChatGPT to perform any mathematical calculations) and are definitively unqualified for evaluation of actual engineering or design quality. These shortcomings indicate that my model is likely evaluating the quality of the inventors through the sophistication of the application text and the self-reported importance of the innovation, rather than scientific inventiveness or importance. Second, GPT-4, along with other "general" developments such as Gato, are the first step towards multimodal models, which will be able to leverage the other parts of a patent, i.e., figures and drawings, which are, in theory, equally important.

**Appendix A. Deep Learning Model Design and Implementation**

I use standard procedure in the layer design, setting node counts as a roughly geometric series (the final layer has no activation, and serves essentially as an OLS regression on the output from the penultimate layer). I utilize 20% dropout regularization to prevent overfitting (the curse of dimensionality, leading to near interpolation of the training sample and poor performance out-of-sample), a process which essentially "drops" 20% of the nodes every epoch of training by setting their parameters to 0. This outperformed other methods of regularization such as L1 and L2, as well as significantly reducing node count and removing regularization. This is likely because it better maintains high-degree interactions within the embedding and avoids the model becoming over-fixated with the individually more impactful structural variables, as could occur with L1 and L2. Finally, the Adam optimizer is used, a process which parametrizes learning rate to obtain the best possible results.

**Appendix B. Example Applications with Low Predicted Chance of Success**

**Title:** Pharmaceutical compositions and methods comprising combinations of 2-alkylidene-19-nor-vitamin D derivatives and parathyroid hormone

**Abstract:** The present invention relates to pharmaceutical compositions and methods of treatment comprising administering to a patient in need thereof a combination of a 2-alkylidene-19-nor-vitamin D derivative and parathyroid hormone or an active fragment or variant thereof. Particularly, the present invention relates to pharmaceutical compositions and methods of treatment comprising administering to a patient in need thereof 2-methylene-19-nor-20(S)-1?,25-dihydroxyvitamin D3 and parathyroid hormone or an active fragment or variant thereof.

**Table 1. Application Success Prediction Results**

This table shows the year-by-year results of the primary predictive model for patent application success, using the machine learning model and the full list of features, including the embedding of the application text. AUC is the area under the Receiver Operating Characteristic (ROC) curve and measures the probability of ranking a positive instance higher than a negative instance. Accuracy is the overall percentage of correct predictions. Precision is the percentage of true positives in all labeled positives, and Recall is the percentage of true positives in all real positives. F1 Score is the harmonic mean of Precision and Recall.

| Year | AUC | F1 Score | Accuracy | Precision | Recall |
|---|---|---|---|---|---|
| 2004 | 60.9% | 85.9% | 77.0% | 79.7% | 93.1% |
| 2005 | 57.9% | 84.8% | 75.1% | 76.3% | 95.3% |
| 2006 | 58.8% | 82.8% | 72.8% | 75.4% | 91.7% |
| 2007 | 58.9% | 81.9% | 71.8% | 74.2% | 91.4% |
| 2008 | 57.6% | 82.8% | 72.5% | 74.9% | 92.5% |
| 2009 | 57.2% | 84.2% | 74.3% | 76.9% | 93.1% |
| 2010 | 58.6% | 84.9% | 75.4% | 80.0% | 90.4% |
| 2011 | 55.6% | 86.9% | 77.6% | 79.7% | 95.6% |
| 2012 | 55.5% | 87.4% | 78.4% | 79.9% | 96.6% |
| 2013 | 56.6% | 86.9% | 77.7% | 79.7% | 95.4% |
| 2014 | 56.4% | 87.3% | 78.4% | 79.8% | 96.5% |
| 2015 | 56.9% | 87.6% | 78.8% | 80.3% | 96.5% |
| 2016 | 55.9% | 88.2% | 79.6% | 80.9% | 97.0% |
| 2017 | 56.2% | 89.4% | 81.4% | 83.2% | 96.6% |
| 2018 | 53.8% | 89.9% | 81.9% | 83.0% | 98.0% |
| 2019 | 55.0% | 88.0% | 79.2% | 81.0% | 96.3% |
| Mean | 57.0% | 86.2% | 77.0% | 79.1% | 94.7% |
| Median | 56.8% | 86.9% | 77.7% | 79.8% | 95.5% |

**Table 2. Application Success Benchmark Comparison**

This table compares the mean and median performance of the full model, an XGBoost classifier trained on the full list of features, and a neural network trained with all features except the embedding of the application text. The variables are defined in Table 1.

|  |  | AUC | F1 Score | Accuracy | Precision | Recall |
|---|---|---|---|---|---|---|
| Full Model | Mean | 57.0% | 86.2% | 77.0% | 79.1% | 94.7% |
|  | Median | 56.8% | 86.9% | 77.7% | 79.8% | 95.5% |
| XGBoost | Mean | 51.4% | 86.5% | 76.5% | 76.8% | 99.0% |
|  | Median | 50.3% | 87.2% | 77.3% | 77.3% | 99.8% |
| No Embedding | Mean | 52.6% | 86.4% | 76.6% | 77.2% | 98.2% |
|  | Median | 52.2% | 87.2% | 77.4% | 77.7% | 98.8% |

**Table 3. Comparison of Best and Worst Predicted Applications**

This table demonstrates the difference in performance between the full model and the benchmark trained without the embedding features when attempting to identify the "best" and "worst" applications. We present the average acceptance rate across all years of the full model and benchmark for the yearly top and bottom 100, 250, 500, and 1000 applications. The top and bottom applications are defined as having the highest and lowest predicted chance of acceptance by the two models, as both output predicted probability of acceptance.

| Yearly Cutoff | Best Success | | | Worst Success | | |
| --- | --- | --- | --- | --- | --- | --- |
| | Full Model | No Embedding | Difference | Full Model | No Embedding | Difference |
| 100 | 96.1% | 87.0% | 9.0% | 26.4% | 37.1% | -10.8% |
| 250 | 95.9% | 86.9% | 9.0% | 27.1% | 38.0% | -11.0% |
| 500 | 95.1% | 86.6% | 8.5% | 29.7% | 40.0% | -10.3% |
| 1000 | 94.1% | 86.7% | 7.4% | 34.5% | 43.5% | -9.0% |

**Table 4. Application Quality and Firm Characteristics**

This table reports the results of the regressions of *application quality* on corporate variables. *Application quality* is proxied through the predicted chance of acceptance by a model trained only on the ChatGPT embedding vector (so that the model only has the textual information as inputs). Firm age is defined as years since the firm was first listed on CRSP. Size is the natural logarithm of shares outstanding times price, adjusted for inflation. Application stock is defined as the cumulative number of all patent applications filed prior to the month of the current application, regardless of success. T-statistics are reported in parentheses, with standard error clustered at the firm level. ***, **, * denote statistical significance at the 0.01, 0.05, and 0.10 levels, respectively.

|  | Application Quality | | | | |
| --- | --- | --- | --- | --- | --- |
|  | (1) | (2) | (3) | (4) | (5) |
| Firm Size | 0.016*** |  |  | 0.013*** | 0.012*** |
|  | (5.37) |  |  | (4.72) | (4.90) |
| Firm Age |  | 0.056*** |  | 0.025** | 0.028*** |
|  |  | (6.73) |  | (2.53) | (3.04) |
| Application Stock |  |  | 0.016*** | 0.011*** | 0.011*** |
|  |  |  | (6.02) | (3.04) | (3.18) |
| Constant | 0.522*** | 0.571*** | 0.637*** | 0.400*** | 0.407*** |
|  | (11.65) | (20.00) | (30.39) | (9.33) | (9.72) |
|  |  |  |  |  |  |
| Patent Controls | No | No | No | No | Yes |
| Firm Fixed Effects | Yes | Yes | Yes | Yes | Yes |
| Observations | 722,163 | 722,163 | 722,163 | 722,163 | 722,163 |
| Adjusted R-squared | 0.265 | 0.268 | 0.268 | 0.271 | 0.296 |

**Table 5. Primary Value Prediction Results**

This table reports the year-by-year results of our prediction model for KPSS valuation, using machine learning and the full list of features, including the embedding of the application text. I report mean squared error, R-squared score, and adjusted R-squared, as well as overall mean and median for all statistics.

| Year | MSE | $R^2$ | Adj. $R^2$ |
|---|---|---|---|
| 2004 | 20.50 | 38.4% | 36.9% |
| 2005 | 22.41 | 40.7% | 39.0% |
| 2006 | 17.89 | 41.0% | 39.6% |
| 2007 | 16.10 | 44.0% | 42.4% |
| 2008 | 18.73 | 42.8% | 41.3% |
| 2009 | 28.15 | 36.3% | 34.7% |
| 2010 | 8.28 | 43.1% | 41.9% |
| 2011 | 10.10 | 40.8% | 39.6% |
| 2012 | 5.61 | 42.1% | 41.0% |
| 2013 | 9.94 | 39.0% | 37.9% |
| 2014 | 8.38 | 39.8% | 38.8% |
| 2015 | 18.93 | 43.5% | 42.6% |
| 2016 | 29.08 | 45.6% | 44.7% |
| 2017 | 36.64 | 48.5% | 47.6% |
| 2018 | 34.25 | 49.1% | 47.9% |
| 2019 | 29.77 | 46.4% | 45.5% |
| 2020 | 43.61 | 48.7% | 47.9% |
| Mean | 21.08 | 42.9% | 41.7% |
| Median | 18.93 | 42.8% | 41.3% |

**Table 6. Patent Value Benchmark Comparison**

This table compares the mean and median performance, averaged across all years, of the full model from Table 5, an XGBoost regressor trained on the full list of features, and a neural network trained with all features except the embedding of the patent text. I report mean squared error, R-squared score, and adjusted R-squared.

|  |  | MSE | $R^2$ | Adj. $R^2$ |
|---|---|---|---|---|
| Full Model | Mean | 21.08 | 42.9% | 41.7% |
|  | Median | 18.93 | 42.8% | 41.3% |
| XGBoost | Mean | 26.87 | 28.5% | 27.0% |
|  | Median | 23.95 | 28.5% | 27.3% |
| No Embedding | Mean | 31.21 | 17.4% | 17.4% |
|  | Median | 27.60 | 17.6% | 17.6% |

## Table 7. Application Screening

This table documents the phenomena of "application screening," in which firms adjust and improve lackluster applications in order to secure their ultimate success. Panel A reports the subsample of applications in which a major change in text is recorded between the application and patent grant (defined as a cosine distance of 0.5), and Panel B the applications with observable changes (defined as a cosine distance of 0.2). The predictive model is run on both the application and patent text, and the improvement is reported. The t-statistics for the mean and median improvements are also reported. ***, **, * denote statistical significance at the 0.01, 0.05, and 0.10 levels, respectively.

Panel A: Cosine distance between patent and application at least 0.05

| | Mean | SD | Min. | 25 Pct. | Median | 75 Pct. | Max. | N |
|---|---|---|---|---|---|---|---|---|
| Predicted Success Rate (Application Text) | 17.4% | 7.7% | 0.7% | 9.8% | 21.4% | 23.3% | 27.4% | 42 |
| Predicted Success Rate (Patent Text) | 27.7% | 19.3% | 3.9% | 13.5% | 23.4% | 35.9% | 83.2% | 42 |
| Improvement | 10.3%*** (3.9432) | 17.0% | -10.0% | -0.5% | 4.1%*** (3.382) | 17.3% | 61.1% | 42 |

Panel B: Cosine distance between patent and application at least 0.02

| | Mean | SD | Min. | 25 Pct. | Median | 75 Pct. | Max. | N |
|---|---|---|---|---|---|---|---|---|
| Predicted Success Rate (Application Text) | 18.8% | 7.9% | 0.7% | 13.5% | 21.9% | 24.1% | 38.9% | 92 |
| Predicted Success Rate (Patent Text) | 25.6% | 15.7% | 3.9% | 13.9% | 23.8% | 30.0% | 83.2% | 92 |
| Improvement | 6.7%*** (4.920) | 13.2% | -10.0% | -0.6% | 1.5%*** (4.381) | 10.2% | 61.1% | 92 |

**Table 8. Comparison of Patent Value Measure with KPSS**

This table reports the deviation of my measure from the KPSS value. I define Adjusted KPSS as the KPSS measure scaled assuming a blanket factor of 72.4% acceptance rate, rather than 55% acceptance, which accounts for some but not all of the difference in valuation. The first two rows show the proportional difference in scaling factor, and the rest shows the comparison between estimated values.

|  | Mean | SD | 10 Pct. | 25 Pct. | Median | 75 Pct. | 90 Pct. | N |
|---|---|---|---|---|---|---|---|---|
| AI Scale/ KPSS | 3.65 | 3.78 | 0.58 | 1.45 | 2.46 | 4.31 | 24.56 | 757560 |
| AI Scale/Adj. KPSS | 2.24 | 2.32 | 0.36 | 0.89 | 1.51 | 2.65 | 15.07 | 757560 |
| AI Value | 30.30 | 70.89 | 0.27 | 1.59 | 8.10 | 25.87 | 72.25 | 481152 |
| KPSS Value | 8.75 | 17.49 | 0.09 | 0.56 | 2.88 | 8.69 | 22.11 | 481152 |
| Adj. KPSS Value | 13.58 | 27.13 | 0.15 | 0.86 | 4.46 | 13.48 | 34.31 | 481152 |
| AI Value - KPSS | 21.55 | 59.35 | 0.03 | 0.43 | 3.79 | 15.81 | 49.87 | 481152 |
| AI Value - Adj. KPSS | 16.72 | 54.35 | -1.44 | 0.04 | 1.77 | 11.18 | 39.78 | 481152 |

**Table 9. Application Strength and Portfolio Returns**

This table reports the monthly performance of the long-short portfolio. The firm-level measure, *application strength*, is defined as the mean predicted chance of acceptance for all applications of a firm published in a given month, multiplied by the square root of the number of applications. The firms are sorted into two groups, above and below median, based on application strength, every month. A long-short portfolio is then constructed for the two groups with a one-month horizon. Abnormal monthly returns are calculated based on the Fama-French three, four, and five factor models. Significance is determined based on the one-tailed t-statistic test. ***, **, * denote statistical significance at the 0.01, 0.05, and 0.10 levels, respectively.

|  | FF3-adjusted Return | FF4-adjusted Return | FF5-adjusted Return |
| --- | --- | --- | --- |
| Low Application Strength | -0.144%* | -0.182%* | -0.151%* |
|  | (-1.30) | (-1.58) | (-1.36) |
| High Application Strength | 0.081%* | 0.094%* | 0.086%* |
|  | (1.31) | (1.50) | (1.31) |
| Difference (High - Low) | 0.235%** | 0.276%** | 0.237%** |
|  | (1.78) | (2.10) | (1.84) |